\documentclass[conference]{IEEEtran}
\IEEEoverridecommandlockouts
\usepackage{cite}
\usepackage{amsmath,amssymb,amsfonts}
\usepackage{amsthm}
\usepackage{algorithmic}
\usepackage{graphicx}
\usepackage{textcomp}
\usepackage{xcolor}
\usepackage{multirow}
\usepackage{booktabs}
\usepackage{algorithm}
\usepackage{algorithmic}
\def\BibTeX{{\rm B\kern-.05em{\sc i\kern-.025em b}\kern-.08em
    T\kern-.1667em\lower.7ex\hbox{E}\kern-.125emX}}

\makeatletter
\newcommand{\linebreakand}{%
  \end{@IEEEauthorhalign}
  \hfill\mbox{}\par
  \mbox{}\hfill\begin{@IEEEauthorhalign}
}
\makeatother

\begin{document}

\title{Poisson-Gamma Dynamical Systems with Time-varying Transition Dynamics\\

}

\author{
  \IEEEauthorblockN{
    Jiahao Wang\textsuperscript{1,\dag}\thanks{\dag\ Equal contribution.}\!,
    Yijun Wang\textsuperscript{2,3,4,\dag},
    Nan Fang\textsuperscript{2},
    Sikun Yang\textsuperscript{2,4,\ddag}\thanks{\ddag\ Corresponding author.}
  }
  \IEEEauthorblockA{
    \textsuperscript{1} Department of Electrical and Computer Engineering, University of Arizona, US \\
    \textsuperscript{2} School of Computing and Information Technology, Great Bay University, China,
    \textsuperscript{3} Tsinghua University, China \\
    \textsuperscript{4} Guangdong Provincial Key Laboratory of Mathematical and Neural Dynamical Systems\\
    Email: jiahaow@arizona.edu,\; wangyj@gbu.edu.cn,\; 2310275052@email.szu.edu.cn,\; sikunyang@gbu.edu.cn
  }
}

\maketitle

\begin{abstract}
Bayesian methodologies for handling count-valued time series have gained prominence due to their ability to infer interpretable latent structures and to estimate uncertainties. Among these Bayesian models, Poisson-Gamma Dynamical Systems (PGDSs) are proven to be effective in capturing the evolving dynamics underlying observed count sequences. However, the state-of-the-art PGDS still falls short in capturing the \emph{time-varying} transition dynamics that are commonly observed in real-world count time series. To mitigate this limitation, a PGDS with time-varying transition kernel (TV-PGDS), is proposed to allow the underlying transition matrices to evolve over time. Three specifically-designed Dirichlet Markov chains (Dir-Dir, Dir-Gam-Dir, PR-Gam-Dir) are constructed to accommodate heterogeneous structural mutations within these dependencies. Leveraging Dirichlet-Multinomial-Beta data augmentation techniques, a fully-conjugate and efficient Gibbs sampler is developed to perform posterior simulation. Experiments show that, in comparison with related models, the proposed PGDS achieves improved predictive performance due to its capacity to learn time-varying dependency structure captured by the time-evolving transition matrices.
\end{abstract}


\linespread{0.96}

\section{Introduction}

There has been an increasing interest in modeling count time series. For instance, some previous works~\cite{blei2006dynamic,wang2006topics,jahnichen2018scalable} are concerned with how to learn the evolving topics behind text corpus (frequencies of words) over time. Some works~\cite{CGMs,raymer2013integrated,wilson2017methods,wanner2021well} try to predict global immigrant trends underlying international population movements. Count time series are often \emph{overdispersed}, \emph{sparse}, \emph{high-dimensional}, and thus can not be well modeled by widely used dynamic models such as linear dynamical systems (LDS)~\cite{kalman1960new,ghahramani1998learning}.  

Recently, many works~\cite{zhou2012augment,zhou2015negative,schein2015bayesian,schein2016poisson,schein2016bayesian,acharya2015nonparametric,schein2019poisson} prefer to adopt non-negative distributions from the gamma-Poisson family to construct hierarchical Bayesian models for count time series. In particular, these models enjoy strong explainability and can estimate uncertainty especially when the observations are \emph{noisy} and \emph{incomplete}. Among these works, Poisson-Gamma Dynamical Systems (PGDSs)~\cite{schein2016poisson} received a lot of attention. PGDS preserves the dynamical mechanism of LDS by evolving the latent state $\theta^{(t)}$ via a gamma Markov chain: $\theta^{(t)}\sim Gam(\tau_{0}\boldsymbol{\Pi}\theta^{(t-1)},\tau_{0})$. Here, the stationary transition kernel $\boldsymbol{\Pi}$ captures complex dynamics by learning how latent dimensions excite one another. For example, a very inspiring research paper may motivate other researchers to publish papers on related topics~\cite{RTMs}. In particular, PGDS can be efficiently learned with a tractable Gibbs sampling scheme via Poisson-Logarithmic data augmentation and marginalization technique~\cite{zhou2015negative}. Due to its strong flexibility, PGDS achieves better performance in predicting missing entities and future observations, compared with related models.

Despite these advantages, although PGDS considered the interdependencies of the latent variables within the Markovian stochastic process, it remains limited by its reliance on static interaction mechanism among these latent variables. This \emph{time-invariant} transition kernel hinders its ability to deeply describe shifting dynamics across time, which are ubiquitous in real-world scenarios ~\cite{WinkelmannBook}. For instance, during the initial stage of the COVID-19 pandemic, the worldwide counts of infectious patients were significantly affected by various local policies, government interventions, and emergent events~\cite{grossman2020political,ihme2021modeling,feng2021impact}. The cross transition dynamics among the different monitoring areas were also evolving as the corresponding policies and interventions changed over time. Hence, PGDS unavoidably makes a certain amount of approximation error in capturing the aforementioned time-varying count time series, using a time-invariant transition kernel.
 
\begin{figure}
    \includegraphics[width=0.9\linewidth]{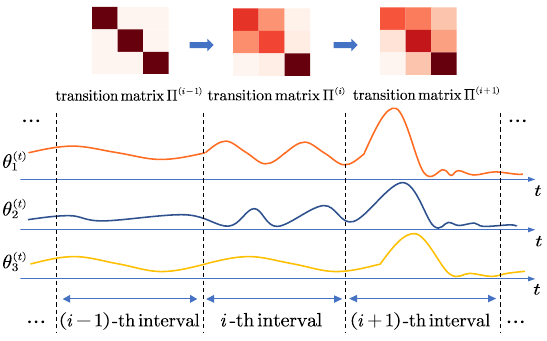}
    \caption{An example illustrates the Poisson-gamma dynamical systems with time-varying transition kernels. The three gamma dynamic processes independently evolve over time during the ($i-1$)-th interval. During $i$-th interval, $\theta_1^{(t)}$ and $\theta_2^{(t)}$ gradually starts to interact with each other while $\theta_3^{(t)}$ remains independent to the other two dimensions. During ($i+1$)-th interval all the three latent components start to interact with each other.}
    \label{fig:diagram}
\end{figure}

To mitigate this limitation, Time-Varying Poisson-Gamma Dynamical
Systems (TV-PGDS), a novel kind of Poisson-gamma dynamical systems with time-varying transition dynamics are developed. In particular, TV-PGDS models the time-varying transition dynamics using tailored Dirichlet Markov chains. Via the Dirichlet-Multinomial-Beta data augmentation strategy, TV-PGDS can be inferred with a conjugate-yet-efficient Gibbs sampler. While ensuring efficient inference, TV-PGDS explicitly captures how does the latent dependencies changes over time as illustrated in Figure \ref{fig:diagram}. As far as we know, this is not realised in existing models. Our contributions are summarized as follows:

1) We propose a time-varying Poisson-Gamma Dynamical System (TV-PGDS), a novel Poisson-gamma dynamical system with time-evolving transition matrices that can well capture time-varying transition dynamics underlying observed count series. 

2) Three Dirichlet Markov chains are dedicated to improving the flexibility and expressiveness of TV-PGDS, for capturing the complex transition dynamics behind sequential count data.

3) Fully-conjugate-yet-efficient Gibbs samplers are developed via Dirichlet-Multinomial-Beta augmentation techniques to perform posterior simulation for the proposed Dirichlet Markov chains.

4) Extensive experiments are conducted on four real-world datasets, to evaluate the performance of the proposed TV-PGDS in predicting missing and future unseen observations. We also provide exploratory analysis to demonstrate the explainable latent structure inferred by the proposed TV-PGDS.

\section{Related Work}

To capture latent dynamics underlying count sequences, previous works ~\cite{han2014dynamic,kalantari2020graph} employed a probabilistic framework $y^{(t)}=p(z^{(t)})$, with $z^{(t)}=f^{-1}(\theta^{(t)})$, where $y^{(t)}\in\mathbb{N}^{V}$ represents count-valued observations at time $t$. The function $p(\cdot)$ is the likelihood, and $f(\cdot)$ maps parameters to continuous latent variables $\theta^{(t)}\in\mathbb{R}^{K}$. In linear dynamical systems (LDS), $\theta^{(t)}$ evolves via a Gaussian Markov chain: $\theta^{(t)}\sim\mathcal{N}(\boldsymbol{\Pi^{(t,t-1)}}\theta^{(t-1)},\boldsymbol{\Lambda^{-1}})$. $\boldsymbol{\Pi}$ is the state transition matrix, and $\boldsymbol{\Lambda}$ is the inverse covariance matrix. While some work ~\cite{han2014dynamic} adopted link functions within the LDS framework to model count observations, it incurs high computational complexity ($\mathcal{O}((K+V)^{3})$), limiting applicability to large-scale sequences. Although LDS and its variants perform well on real-valued sequences, their Gaussian state-space assumption hinders inference efficiency and fails to accurately model burstiness in count time series. 
Recently, Poisson-gamma family models~\cite{acharya2015nonparametric,schein2016poisson,schein2019poisson} have emerged for count sequences. Gamma Process Dynamic Poisson Factor Analysis (GP-DPFA)~\cite{acharya2015nonparametric} models count data as $y_v^{(t)} \sim \mathrm{Pois} ( \sum_{k=1}^K \lambda_k \phi_{vk} \theta_k^{(t)} )$, where $\theta_k^{(t)}$ represents the strength of $k$-th latent factor at time $t$, and $\phi_{vk}$ captures the involvement degree of $k$-th factor to $v$-th observed dimension, regularized by a Dirichlet prior $\phi_{k}\sim Dir(\epsilon_{0},\cdot\cdot\cdot,\epsilon_{0})$. The latent factor evolves via a gamma Markov chain: ${\theta}_{k}^{(t)}\sim Gam(\theta_{k}^{(t-1)},c_{t})$. Although GP-DPFA fits one-dimensional sequences well, it fails to learn interactions among latent dimensions.   
To address this concern, Poisson-Gamma Dynamical Systems (PGDS)~\cite{schein2016poisson} captures underlying transition dynamics by evolving $\theta_k^{(t)}$ as $\theta_k^{(t)} \sim \mathrm{Gam} ( \tau_0 \sum_{k_2 = 1}^K \pi_{kk_2} \theta_{k_2}^{(t-1)}, \tau_0 )$, where $\pi_{kk_2}$ represents how $k_2$-th latent factor excites the $k$-th latent factor at next time step, and $\sum_{k=1}^K \pi_{kk_2} = 1$. Several extensions of PGDS are developed to further enhance the flexibility of the gamma Markov chain in capturing dynamics. Switching PGDS (SPGDS) ~\cite{chen2021switching} models non-linear dynamics via discrete switching among finite transition regimes as $\theta_{k}^{(t)}\sim\prod_{c=1}^{C}Gam(\tau_{0}\Pi_{c}\theta_{k_{2}}^{(t-1)},\tau_{c})$. However, they are constrained by stationary assumptions—DPGDS employs fixed kernels, while SPGDS aggregates discrete states—making them inadequate for continuously evolving non-stationary environments. 
 
Another stream of studies models dynamics by chaining the rate parameter of the gamma distribution. For example, GMC-RATE ~\cite{fevotte2009nonnegative} formulates $\theta_{k}^{(t)}\sim \mathrm{Gam}(\alpha,\beta/\theta_{k}^{(t-1)})$. GMC-HIER ~\cite{filstroff2021comparative} introduces hierarchical auxiliary variables, such as $z_{k}^{(t)}\sim \mathrm{Gam}(\alpha_{z},\beta_{z}\theta_{k}^{(t-1)})$ and $\theta_{k}^{(t)}\sim \mathrm{Gam}(a_{\theta},\beta_{\theta}z_{k}^{(t)})$. However, since the rate parameter governs variance, these models are prone to high volatility, potentially causing numerical instability or model collapse.  

\section{Time-Varying Poisson-Gamma Dynamical Systems}

Real-world count time sequences are often \emph{time-varying} because the external interventional environments are always changing over time. The stationary PGDS with a time-invariant transition kernel fails to capture such time-varying transition dynamics. Hence, to mitigate this limitation, we model the count sequences as
\begin{equation}
    y_v^{(t)} \sim \mathrm{Pois} \left( \delta^{(t)} \textstyle \sum_{k=1}^{K} \phi_{vk} \theta_k^{(t)} \right), \nonumber
\end{equation}
in which, the latent factors are specified by
\begin{equation}
    \theta_k^{(t)} \sim \mathrm{Gam} \left( \tau_0 \textstyle \sum_{k_2 = 1}^K \pi_{kk_2}^{(t-1)} \theta_{k_2}^{(t-1)}, \tau_0 \right) \label{ns-pgds},
\end{equation}
where the multiplicative term $\delta^{(t)} \sim \mathrm{Gam} \left( \epsilon_0, \epsilon_0 \right)$ and the transition matrices are time-varying as $\boldsymbol{\Pi}^{(t)} \equiv [ \pi_{kk_2}^{(t)} ]_{k, k_2=1}^K$.

\begin{figure*}[t]
    \centering
    \includegraphics[width=1.0\linewidth]{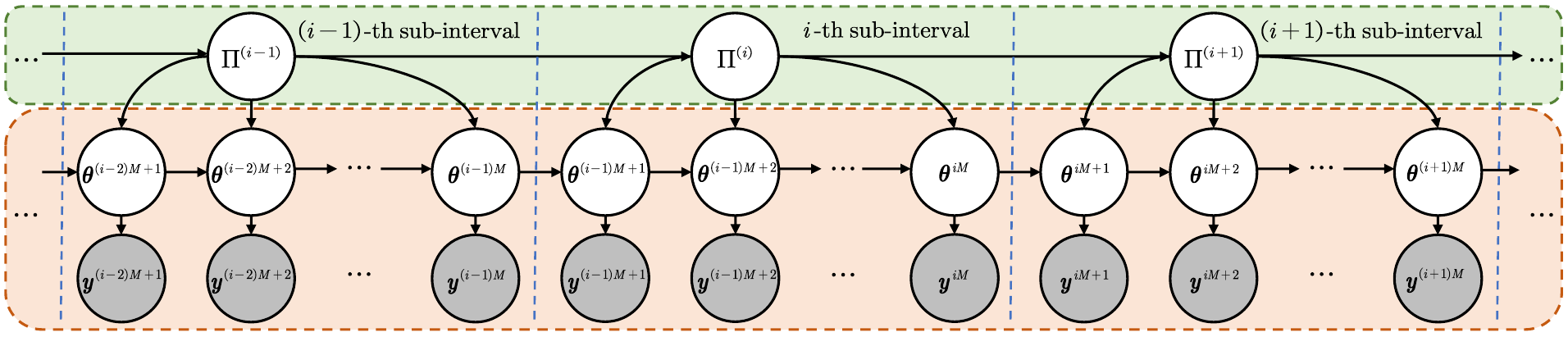}
    \caption{The graphical representation of the TV-PGDS. The time interval is divided into equally-spaced sub-intervals. Each sub-interval contains $M$ time steps. The transition dynamics is stationary within a sub-interval. In particular, the transition matrices evolve over sub-intervals via Dirichlet Markov processes while latent factors evolve over time steps via Eq.(\ref{ns-pgds}).}
    \label{fig:graphical_rep}
\end{figure*}

As shown in Figure \ref{fig:diagram}, to model the time-varying transition dynamics, we assume the whole time interval can be divided into $I$ equally-spaced sub-intervals. The transition kernel behind complicated dynamic counts is assumed to be \emph{static} within each sub-interval, while evolving over sub-intervals, to capture time-varying behaviours. In other words, the proposed model allows the latent factors to evolve over time steps while the transition matrices change over sub-intervals but assumed to be stationary within each sub-interval, as shown in Figure \ref{fig:graphical_rep}. In particular, we let each sub-interval contains $M$ time steps, and the $i$-th interval contains time steps $\left\{ t \mid t=\left( i-1\right)M+1, \cdots, iM \right\}$. We define $i\left(t \right)$ as the function that maps time step $t$ to its corresponding sub-interval.

By setting local stationarity within each sub-interval, the transition matrices mitigate the overfitting risks inherent in high-dimensional temporal modeling. The sub-interval length, $M$, governs a fundamental trade-off: while $M=1$ characterizes a fully dynamic system with maximum granularity, such a configuration often leads to high computational costs or unstable convergence. Hence, the value of $M$ should be determined by the patterns of empirical data and model performance. An analysis of the influence of interval length $M$ on model performance is provided in the following section.

\noindent \textbf{Dirichlet-Dirichlet Markov processes.} To capture how the underlying transition kernel smoothly evolves over sub-intervals, we first propose the Dirichlet-Dirichlet (Dir-Dir) Markov chain as
\begin{equation}
\label{chain1}
    \boldsymbol{\pi}_k^{(i)} \mid \boldsymbol{\pi}_k^{(i-1)} \sim \mathrm{Dir} \left( \eta K \pi_{1k}^{(i-1)}, \cdots, \eta K \pi_{Kk}^{(i-1)} \right),
\end{equation}
where $\boldsymbol{\pi}_k^{(i)}$ represents the $k$-th column of $\boldsymbol{\Pi}^{(i)}$, and the prior of the scaling parameter $\eta$ is given by $\eta \sim \mathrm{Gam} \left( e_0, f_0 \right)$.

The initial states are defined as $\theta_k^{(1)} \sim \mathrm{Gam} \left( \tau_0 \nu_k, \tau_0 \right)$. The prior for the transition kernel of the first sub-interval is given by $\boldsymbol{\pi}_k^{(1)} \sim \mathrm{Dir} \left( \nu_1 \nu_k, \cdots, \xi \nu_k, \cdots, \nu_K \nu_k \right)$, where $\nu_k \sim \mathrm{Gam} ( \frac{\gamma_0}{K}, \beta ) \; \mathrm{and} \; \xi, \beta \sim \mathrm{Gam} \left( \epsilon_0, \epsilon_0\right)$. $\nu_k$ captures the global prominence/activity of latent factor $k$, while $\nu_{k_1}\nu_{k_2}$ provides a separable prior preference for transitions $k_1\to k_2$. Note that the expectation and variance of the transition kernel at $i$-th sub-interval can be calculated as
\begin{align}
    \mathsf{E}\left[ \boldsymbol{\pi}_k^{(i)} \mid \boldsymbol{\pi}_k^{(i-1)} \right] &= \boldsymbol{\pi}_k^{(i-1)}, \nonumber \\
    \mathsf{Var} \left[ \pi_{k_1k}^{(i)} \mid \boldsymbol{\pi}_k^{(i-1)} \right] &= \frac{\pi_{k_1k}^{(i-1)} \left( 1 - \pi_{k_1k}^{(i-1)} \right)}{\eta K + 1} \nonumber,
\end{align}
respectively. The transition dynamics of  $i$-th sub-interval inherits the information of the previous sub-interval, and also adapts to the data observed in the current sub-interval. The parameter $\eta$ controls the variance of the transition matrices.

The prior specification defined in Eq.(\ref{chain1}) by rescaling the transition matrix at the previous sub-interval allows the transition dynamics to change smoothly, and thus might be insufficient to capture the rapid mutations observed in complicated dynamics.  To further improve the flexibility of the transition structure, two modified Dirichlet Markov chains are studied to capture the correlation structure between the dimensions of the transition matrices over time. 
\begin{figure}
    \includegraphics[width=0.9\linewidth]{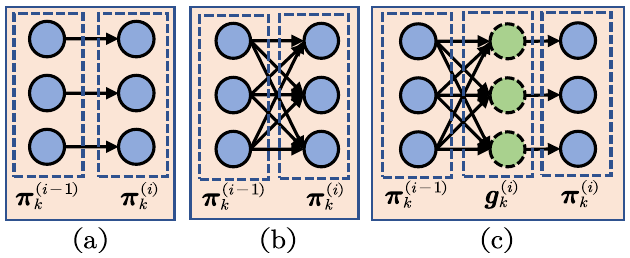}
    \centering
    \caption{Diagrams of the proposed Dirichlet Markov constructions. (a) is the Dir-Dir construction. (b) is the Dir-Gam-Dir construction which takes mutation into account. (c) illustrates the PR-Gam-Dir construction which adopts Poisson randomized gamma distribution and can be equivalently represented as Eq.(\ref{pr_gam_chain_1}).}
    \label{fig:three_chains}
\end{figure}

\noindent \textbf{Dirichlet-Gamma-Dirichlet Markov processes.} We first introduce the Dirichlet-Gamma-Dirichlet (Dir-Gam-Dir) Markov chain to model the evolving transition matrices as
\begin{align}
    \boldsymbol{\pi}_k^{(i)} &\sim \mathrm{Dir} \left( \alpha_{1k}^{(i)}, \cdots, \alpha_{Kk}^{(i)} \right), \\ \alpha_{k_1k}^{(i)} &\sim \mathrm{Gam} \left( \gamma_k^{(i-1)} \textstyle \sum_{k_2=1}^K \psi^{(i-1)}_{kk_1k_2}\pi_{k_2k}^{(i-1)}, c_k^{(i)} \right), \nonumber
\end{align}
where we use $\psi_{kk_1k_2}^{(i-1)}$ to capture the mutation between two consecutive sub-intervals, and its prior is given by
\begin{align}
    \left(\psi_{k1k_2}^{(i-1)}, \cdots, \psi_{kKk_2}^{(i-1)} \right) \sim \mathrm{Dir} \left( \epsilon_0, \cdots, \epsilon_0\right), 
\end{align}
and $\gamma_k^{(i)}, c_k^{(i)} \sim \mathrm{Gam}\left( \epsilon_0, \epsilon_0\right)$. Compared with the construction defined by Eq.(\ref{chain1}), the expectation of Dirichlet-Gamma-Dirichlet Markov chain is
\begin{equation}
    \mathsf{E}\left[ \boldsymbol{\pi}_k^{(i)} \mid \boldsymbol{\pi}_k^{(i-1)}\right] = \boldsymbol{\Psi}_k^{(i-1)} \boldsymbol{\pi}_k^{(i-1)}. \nonumber
\end{equation}
This construction takes interactions among components of columns into account. Hence it will improve the flexibility of our model and thus better fit more complicated dynamics, compared with Dir-Dir Markov chains that only yield smoothing transition dynamics. 

\noindent \textbf{Poisson-randomized-gamma-Dirichlet Markov processes.} By leveraging the Poisson-randomized gamma distribution~\cite{yuan2000bessel}, we introduce another type of time-varying transition kernels, which also model the interactions among components like Dir-Gam-Dir construction but may further encode a unique inductive bias tailored to sparsity and burstiness. The Poisson-randomized-gamma-Dirichlet (PR-Gam-Dir) Markov chain can be formulated as
\begin{align}
\label{pr_gam_chain}
    \boldsymbol{\pi}_k^{(i)} &\sim \mathrm{Dir} \left( \alpha_{1k}^{(i)}, \cdots, \alpha_{Kk}^{(i)} \right), \\ 
    \alpha_{k_1k}^{(i)} &\sim \mathrm{RG1} \left( \epsilon^\alpha, \gamma_k^{(i-1)} \textstyle \sum_{k2=1}^K \psi^{(i-1)}_{kk_1k_2}\pi_{k_2k}^{(i-1)}, c_k^{(i)}\right), \nonumber 
\end{align}
where $\mathrm{RG1} \left( \cdot \right)$ denotes the randomized gamma distribution of the first type. Similarly, for $\psi_{kk_1k_2}^{(i-1)}$, $\gamma_k^{(i)}$, and $c_k^{(i)}$, the priors are given by
\begin{align}
    \left(\psi_{k1k_2}^{(i-1)}, \cdots, \psi_{kKk_2}^{(i-1)} \right) &\sim \mathrm{Dir} \left( \epsilon_0, \cdots, \epsilon_0\right), \\
    \gamma_k^{(i)}, c_k^{(i)} &\sim \mathrm{Gam}\left( \epsilon_0, \epsilon_0\right), \, \mathrm{respectively}. \nonumber
\end{align}

The PR-Gam-Dir Markov chain in Eq.(\ref{pr_gam_chain}) can be equivalently written as 
\begin{align}
\label{pr_gam_chain_1}
    \boldsymbol{\pi}_k^{(i)} &\sim \mathrm{Dir} \left( \alpha_{1k}^{(i)}, \cdots, \alpha_{Kk}^{(i)} \right),  \nonumber \\ 
    \alpha_{k_1k}^{(i)} &\sim \mathrm{Gam} \left( g_{k_1k}^{(i)} + \epsilon^\alpha, c_k^{(i)} \right),\\
    g_{k_1k}^{(i)} &\sim \mathrm{Pois} \left( \gamma_k^{(i-1)} \textstyle \sum_{k2=1}^K \psi^{(i-1)}_{kk_1k_2}\pi_{k_2k}^{(i-1)}\right). \nonumber 
\end{align}
When we marginalize out $g_{k_1k}^{(i)}$ in Eq.(\ref{pr_gam_chain_1}), we then obtain $\alpha_{k_1k}^{(i)} \sim \mathrm{RG1} \left( \epsilon^\alpha, \gamma_k^{(i-1)} \textstyle \sum_{k2=1}^K \psi^{(i-1)}_{kk_1k_2}\pi_{k_2k}^{(i-1)}, c_k^{(i)}\right)$ in Eq.(\ref{pr_gam_chain}). Compared with gamma distribution, RG1 enables a deeper decomposition of complex dynamics ~\cite{schein2019poisson}. The Dir-Dir and Dir-Gam-Dir Markov chains were restricted by similar patterns between transitional matrices. The Poisson latent variable $\boldsymbol{g}_{k}^{(i)}$ in the PR-Gam-Dir Markov chain decouples the component ratios, further enhancing the model's representation capability. Thus, this modification may allow for large mutations in latent distributions and accommodate to burstiness. 

The diagrams of three Dirichlet Markov constructions are shown in Figure \ref{fig:three_chains}.

\section{Markov Chain Monte Carlo Inference}

In this section, we present the Gibbs sampler for the proposed TV-PGDS. We only illustrate the key points of the derivation and the details can be found in the appendix. In the following inference, we repeatedly used two lemmas.
\newtheoremstyle{boldlemma} 
  {3pt}            
  {3pt}               
  {\itshape}         
  {0pt}             
  {\bfseries}       
  {.}                 
  {.5em}              
  {}                   

\theoremstyle{boldlemma}
\newtheorem{lemma}{Lemma}

\begin{lemma}
\label{poisson-crt}
    If $y \sim \mathrm{NB} \left( a, g \left( \zeta \right) \right)$ and $l \sim \mathrm{CRT} \left( y, a\right)$, where $\mathrm{NB} \left( \cdot \right)$ refers to negative-binomial distribution, $\mathrm{CRT} \left( \cdot \right)$ represents Chinese restaurant table distribution~\cite{YWT2006hier}, and $g\left( z \right) = 1 - \mathrm{exp} \left( -z \right)$. Then the joint distribution of $y$ and $l$ can be equivalently distributed as $y \sim \mathrm{SumLog} \left( l, g \left( \zeta \right) \right)$ and $l \sim \mathrm{Pois} \left( a \zeta \right)$~\cite{zhou2015negative}, i.e.
    $$
    \begin{aligned}
        &\mathrm{NB} \left(y; a, g \left( \zeta \right) \right) \mathrm{CRT} \left(l; y, a\right) =\\
        &\mathrm{SumLog} \left(y; l, g \left( \zeta \right) \right) \mathrm{Pois} \left(l; a \zeta \right),
    \end{aligned}
    $$
    where $\mathrm{SumLog} \left(l, g \left( \zeta \right) \right) = \sum_{i=1}^l x_i$ and $x_i \sim \mathrm{Log} \left( 
    g \left( \zeta \right) \right)$ are independently and identically logarithmic distributed random variables~\cite{johnson2005univariate}.
\end{lemma}
\begin{lemma}
\label{dirmult}
    Suppose $\mathbf{n} = \left( n_1, \cdots, n_K\right)$ and
    $$
        \mathbf{n} \mid n \sim \mathrm{DirMult} \left( n, r_1, \cdots, r_K \right),
    $$
    where $\mathrm{DirMult} \left( \cdot \right)$ refers to Dirichlet-multimonial distribution. We sample the augmented variable $q \mid n \sim \mathrm{Beta}\left( n, r_{\cdot}\right)$,
    where $r_{\cdot} = \sum_{k=1}^K r_k$. According to~\cite{zhou2018nonparametric}, conditioning on $q$, we have $n_k \sim \mathrm{NB} \left( r_k, q\right)$.
\end{lemma}

\noindent \textbf{Sampling $y_{vk}^{(t)}$ :} Use the relationship between Poisson and multinomial distributions, we sample
\begin{equation}\label{sampling_y}
\resizebox{0.88\linewidth}{!}{$
    \begin{aligned}
    \left( \left( y_{vk}^{(t)} \right)_{k=1}^K \mid - \right) \sim \mathrm{Mult} \left( y_v^{(t)}, \left( \frac{\phi_{vk} \theta_k^{(t)}}{\sum_{k=1}^K \phi_{vk} \theta_k^{(t)}} \right)_{k=1}^K \right).
    \end{aligned}
$}
\end{equation}
\noindent \textbf{Sampling $\boldsymbol{\phi_k}$ :} Via Dirichlet-multinomial conjugacy, the posterior of $\boldsymbol{\phi_k}$ is
\begin{equation}\label{sampling_phi}
\resizebox{0.88\linewidth}{!}{$
    \begin{aligned}
    \left( \boldsymbol{\phi_k} \mid -\right) \sim \mathrm{Dir} \left( \epsilon_0 + \textstyle \sum_{t=1}^T y_{1k}^{(t)}, \cdots, \epsilon_0 + \textstyle \sum_{t=1}^T y_{Vk}^{(t)}\right).
    \end{aligned}
$}
\end{equation}
\noindent \textbf{Sampling $\theta_k^{(t)}$ :} To sample from the posterior of $\theta_k^{(t)}$, we first sample the auxiliary variables. The auxiliary variables $l_{k\cdot}^{(t)}$ are used to decouple $\theta_k^{(t)}$ from the Markov chain. The backward decoupling process is detailed in the appendix. Setting $l_{\cdot k}^{(T+1)} = 0$ and $\zeta^{(T+1)} = 0$, we sample the augmented variables backwards from $t = T, \cdots, 2$,
\begin{equation}\label{sampling_l}    
\resizebox{0.88\linewidth}{!}{$
    \begin{aligned}
    &\left( l_{k\cdot}^{(t)} \mid - \right) \sim \mathrm{CRT} \left( y_{\cdot k}^{(t)} + l_{\cdot k}^{(t+1)}, \tau_0 \textstyle \sum_{k_2=1}^K \pi_{kk_2}^{i\left( t-1 \right)} \theta_{k_2}^{(t-1)} \right),\\
    &\left( l_{k1}^{(t)}, \cdots, l_{kK}^{(t)} \mid - \right) \sim \\
    &\mathrm{Mult} \left( l_{k\cdot}^{(t)}, \left(\frac{\pi_{k1}^{i\left( t-1 \right)} \theta_1^{(t-1)}}{\sum_{k_2=1}^{K} \pi_{kk_2}^{i\left( t-1 \right)} \theta_{k_2}^{(t-1)} }, \cdots, \frac{\pi_{kK}^{i\left( t-1 \right)} \theta_K^{(t-1)}}{\sum_{k_2=1}^{K} \pi_{kk_2}^{i\left( t-1 \right)} \theta_{k_2}^{(t-1)} } \right) \right).
    \end{aligned}
$}
\end{equation}
Let us define $l_{\cdot k}^{(t)} = \sum_{k_1=1}^K l_{k_1k}^{(t)}$ and $\zeta^{(t)} = \mathrm{ln} ( 1 + \frac{\delta^{(t)}}{\tau_0} + \zeta^{(t+1)} )$. Using \emph{Lemma \ref{poisson-crt}}, $l_{k\cdot}^{(t)}$ can also be written in 
\begin{equation}\label{lk_poisson}
l_{\cdot k}^{(t)} \sim \mathrm{Pois} \left(\zeta^{(t)}\tau_0\theta_k^{(t-1)}\right).
\end{equation}

After sampling the auxiliary variables, for $t=1, \cdots, T$, by Poisson-gamma conjugacy, we obtain
\begin{equation}\label{sampling_theta}
\resizebox{0.88\linewidth}{!}{$
    \begin{aligned}
    \left( \theta_k^{(t)} \mid - \right) \sim \mathrm{Gam} &(y_{\cdot k}^{(t)} + l_{\cdot k}^{(t+1)} + \tau_0 \textstyle \sum_{k_2=1}^K \pi_{kk_2}^{i\left( t-1\right)} \theta_{k_2}^{(t-1)}, \\
    &\tau_0 + \delta^{(t)} + \zeta^{(t+1)} \tau_0 ).
\end{aligned}
$}
\end{equation}

\begin{figure}
    \centering
    \includegraphics[width=5.5 cm]{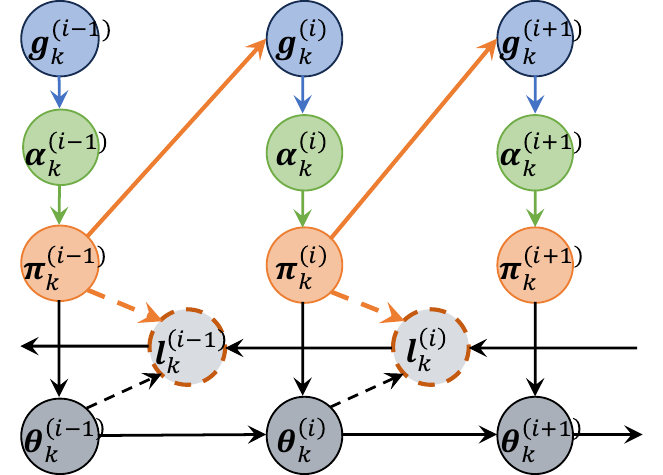}
    \caption{The probabilistic graphical model of the proposed PR-Gam-Dir construction as Eq.(\ref{pr_gam_chain_1}). Dashed nodes and edges denote the augmented auxiliary variables.}
    \label{fig:pr_gm_dir_rep}
\end{figure}

\noindent \textbf{Sampling $\boldsymbol{\Pi}^{(i)}$ :} We only illustrate Gibbs sampling algorithm for PR-Gam-Dir construction. Sampling algorithms for other constructions can be found in the appendix. The probabilistic graphical model of PR-GM-Dir construction is shown in Figure \ref{fig:pr_gm_dir_rep}. We define $M$ denote the length of each sub-interval and $I$ be the total number of intervals. For $i \in \{2, \dots, I\}$ and $k_1 \in \{1, \dots, K\}$, $l_{k_1k}^{(i)}$ and $g_{\cdot k_1k}^{(i+1)}$ are characterized by Poisson distributions, as formulated in Eq.(\ref{pr_gam_chain_1}) and Eq.(\ref{lk_poisson}). Here, $l_{k_1k}^{(i)} = \sum_{\left( i-1 \right)M+1}^{iM}l_{k_1k}^{(t)}$ refers to the summation of $l_{k_1k}^{(t)}$ over $i$-th sub-interval; analogous notation applies to other variables. By exploiting the conjugacy between Poisson and multinomial distributions, the vectors $( l_{1k}^{(i)}, \cdots, l_{Kk}^{(i)})$ and $( g_{\cdot 1k}^{(i+1)},\cdots, g_{\cdot Kk}^{(i+1)} )$ are multinomially distributed with parameter $( \pi_{1k}^{(i)}, \cdots, \pi_{Kk}^{(i)})$, as in Figure \ref{fig:pr_gm_dir_rep}.  Upon marginalizing out $\boldsymbol{\pi}_{k}^{(i)}$, $l_{\cdot k}^{(i)}$ and $g_{\cdot k}^{(i+1)}$ follow Dirichlet-Multinomial distributions.

Consequently, we construct an auxiliary variable $q_k^{(i)}$ via \emph{Lemma \ref{dirmult}}, defining $g_{k_1k}^{(I+1)}=0$, as 
\begin{equation}\label{sampling_q}
( q_k^{(i)} \mid -) \sim \mathrm{Beta} ( l_{\cdot k}^{(i)}+g_{\cdot k}^{(i+1)}, \alpha_{\cdot k}^{(i)}).
\end{equation}

Then we have $( l_{k_1k}^{(i)} + g_{\cdot k_1k}^{(i+1)} ) \sim \mathrm{NB}(\alpha_{k_1k}^{(i)}, q_k^{(i)} )$. Furthermore, by applying the NB-CRT augmentation (\emph{Lemma \ref{poisson-crt}}), we sample another auxiliary variable $h_{k_1k}^{(i)}$ as
\begin{equation}\label{sampling_h}
(h_{k_1k}^{(i)} \mid - ) \sim \mathrm{CRT} ( l_{k_1k}^{(i)}+g_{\cdot k_1k}^{(i+1)}, \alpha_{k_1k}^{(i)}),
\end{equation}
which, according to the properties of the CRT distribution, yields $h_{k_1k}^{(i)} \sim \mathrm{Pois} ( -\alpha_{k_1k}^{(i)} \mathrm{ln} ( 1 - q_k^{(i)} ))$. By Poisson-gamma conjugacy, we have $( \alpha_{k_1k}^{(i)} \mid - ) \sim \mathrm{Gam} ( g_{k_1k}^{(i)} + \epsilon^{\alpha} + h_{k_1k}^{(i)}, c_k^{(i)}-\mathrm{ln}( 1-q_k^{(i)}))$. 

We define $\lambda_{k_1k}^{(i-1)} \triangleq \gamma_k^{(i-1)}\sum_{k2=1}^K \psi^{(i-1)}_{kk_1k_2}\pi_{k_2k}^{(i-1)}$ for notation conciseness. When sampling $g_{k_1k}^{(i)}$, the posterior distribution—arising from the Poisson prior and the gamma likelihood—exhibits a closed-form analytical solution via the Bessel distribution \cite{yuan2000bessel}. Specifically, if $\epsilon^\alpha > 0$, then we can sample the posterior of $g_{k_1k}^{(i)}$ via $( g_{k_1k}^{(i)} \mid - ) \sim \mathrm{Bessel} ( \epsilon^\alpha-1, 2 \sqrt{\alpha_{k_1k}^{(i)}c_k^{(i)} \lambda_{k_1k}^{(i-1)}}\,)$, where $\mathrm{Bessel} \left( \cdot \right)$ denotes Bessel distribution. If $\epsilon^\alpha = 0$, to avoid the absorbing condition between $\alpha_{k_1k}^{(i)}$ and $g_{k_1k}^{(i)}$, we sample $g_{k_1k}^{(i)}$ via
\begin{equation}\label{sampling_g}
\resizebox{0.88\linewidth}{!}{ 
    $\displaystyle 
    (g_{k_1k}^{(i)} \mid -) \sim
    \begin{cases} 
        \mathrm{Pois} \left( \frac{c_k^{(i)} \lambda_{k_1k}^{(i-1)}}{c_k^{(i)}-\ln(1 - q_k^{(i)})} \right) & \text{if } h_{k_1k}^{(i)}=0 \\
        \mathrm{SCH} \left( h_{k_1k}^{(i)}, \frac{c_k^{(i)} \lambda_{k_1k}^{(i-1)}}{c_k^{(i)}-\ln(1 - q_k^{(i)})} \right) & \text{otherwise},
    \end{cases}
    $
}
\end{equation}
where $\mathrm{SCH} \left( \cdot \right)$ denotes the shifted confluent hypergeometric distribution~\cite{schein2019poisson}. Both Bessel and the shifted confluent hypergeometric distribution can be sampled efficiently \cite{devroye2002simulating,schein2019poisson}. Defining $g_{k_1k}^{(i)}=g_{k_1\cdot k}^{(i)}=\sum_{k2=1}^K g_{k_1k_2k}^{(i)}$, we first augment
\begin{equation}\label{sampling_gg}\small
    \left( g_{k_11k}^{(i)}, \cdots, g_{k_1Kk}^{(i)}\right) \sim \mathrm{Mult} \Big(g_{k_1k}^{(i)},\big( \psi^{(i-1)}_{kk_1k_2}\pi_{k_2k}^{(i-1)} \big)_{k_2=1}^K \Big).
\end{equation}
Then we obtain $g_{k_1k_2k}^{(i)} \sim \mathrm{Pois}(\gamma^{(i-1)} \psi^{(i-1)}_{kk_1k_2}\pi_{k_2k}^{(i-1)})$. By Dirichlet-multinomial conjugacy, we have
\begin{align}
    &\Big( \big( \psi_{k1k_2}^{(i-1)}, \dots, \psi_{kKk_2}^{(i-1)} \big) \mid - \Big) \sim \mathrm{Dir} \Big( \epsilon_0 + g_{1k_2k}^{(i)}, \dots, \nonumber \\
    & \qquad \qquad \qquad \qquad\qquad \epsilon_0 + g_{Kk_2k}^{(i)} \Big), \label{sampling_psi} \\
    & \left( \boldsymbol{\pi}_k^{(i-1)} \mid - \right) \sim \mathrm{Dir} \Big( \alpha_{1k}^{(i-1)}+l_{1k}^{(i-1)}+g_{\cdot 1k}^{(i)}, \dots, \nonumber \\ 
    & \qquad \qquad \qquad \alpha_{Kk}^{(i-1)}+l_{Kk}^{(i-1)}+g_{\cdot Kk}^{(i)} \Big). \label{sampling_pi}
\end{align}

Specifically, we have $\alpha_{k_1k}^{(1)}=\nu_{k_1}\nu_k$, if $k_1 \ne k$, and $\alpha_{k_1k}^{(1)}=\xi \nu_k$, if $k_1=k$.

\section{Experiments}

\begin{table*}[htbp]
\caption{Results of predictive analysis. "S" means data smoothing and "F" means data forecasting.}
\label{tab:predictive_results}
\vspace{-1em}
    \centering
    \tabcolsep=1mm
    \resizebox{\textwidth}{!}{
    \begin{tabular}{cccrrrrrrrr}
    \toprule[2pt]
         \multirow{2}{*}{} & \multirow{2}{*}{} & \multirow{2}{*}{} & \multicolumn{1}{c}{\multirow{2}{*}{GP-DPFA}} & \multicolumn{1}{c}{\multirow{2}{*}{GMC-RATE}} & \multicolumn{1}{c}{\multirow{2}{*}{GMC-HIER}} & \multicolumn{1}{c}{\multirow{2}{*}{BGAR}} & \multicolumn{1}{c}{\multirow{2}{*}{PGDS}} & \multicolumn{1}{c}{TV-PGDS} & \multicolumn{1}{c}{TV-PGDS} & \multicolumn{1}{c}{TV-PGDS} \\ 
        \multicolumn{8}{c}{} & \multicolumn{1}{c}{(Dir-Dir)} & \multicolumn{1}{c}{(Dir-Gam-Dir)} & \multicolumn{1}{c}{(PR-Gam-Dir)}\\ 
        \midrule
        \multirow{4}{*}{} ICEWS & \multirow{2}{*}{} MAE & S & 0.259 \scriptsize $\pm 0.005$ & 0.258 \scriptsize $\pm 0.005$& 0.256 \scriptsize $\pm 0.006$ & 0.264 \scriptsize $\pm 0.006$ & 0.215 \scriptsize $\pm 0.007$ & 0.215 \scriptsize $\pm 0.008$ & \textbf{0.214} \scriptsize $\pm 0.008$ & 0.215 \scriptsize $\pm 0.008$\\
        
        & & F & 0.176 \scriptsize $\pm 0.005$ & 0.187 \scriptsize $\pm 0.003$ & 0.185 \scriptsize $\pm 0.016$ & 0.222 \scriptsize $\pm 0.043$ & 0.185 \scriptsize $\pm 0.003$ & \textbf{0.167} \scriptsize $\pm 0.009$ & 0.169 \scriptsize $\pm 0.006$ & 0.169 \scriptsize $\pm 0.009$\\
        
        & \multirow{2}{*}{} MRE & S & 0.125 \scriptsize $\pm 0.003$ & 0.124 \scriptsize $\pm 0.002$ & 0.122 \scriptsize $\pm 0.003$ & 0.130 \scriptsize $\pm 0.004$ & 0.102 \scriptsize $\pm 0.005$ & \textbf{0.101} \scriptsize $\pm 0.005$ & \textbf{0.101} \scriptsize $\pm 0.005$ & 0.102 \scriptsize $\pm 0.005$\\
        
        & & F & 0.099 \scriptsize $\pm 0.006$ & 0.114 \scriptsize $\pm 0.003$ & 0.111 \scriptsize $\pm 0.018$ & 0.142 \scriptsize $\pm 0.036$& 0.108 \scriptsize $\pm 0.001$ & \textbf{0.094} \scriptsize $\pm 0.005$ & 0.097 \scriptsize $\pm 0.004$ & 0.097 \scriptsize $\pm 0.008$\\
        \midrule
        
        \multirow{4}{*}{} NIPS & \multirow{2}{*}{} MAE & S & 18.299 \scriptsize $\pm 6.545$ & 17.105 \scriptsize $\pm 6.449$ & 17.098 \scriptsize $\pm 6.441$ & 17.935 \scriptsize $\pm 6.450$ & 14.706 \scriptsize $\pm 4.414$ & 14.032 \scriptsize $\pm 4.401$ & 14.026 \scriptsize $\pm 4.405$ & \textbf{14.014} \scriptsize $\pm 4.387$\\
        
        & & F & 48.355 \scriptsize $\pm 1.461$ & 46.234 \scriptsize $\pm 1.629$ & 102.506 \scriptsize $\pm 39.932$ & 62.449 \scriptsize $\pm 14.463$ & 51.562 \scriptsize $\pm 0.679$ & \textbf{45.979} \scriptsize $\pm 1.342$ & 46.710 \scriptsize $\pm 1.152$ & 46.582 \scriptsize $\pm 1.196$\\
        
        & \multirow{2}{*}{} MRE & S & 0.729 \scriptsize $\pm 0.412$ & 0.684 \scriptsize $\pm 0.316$ & 0.664 \scriptsize $\pm 0.315$ & 0.769 \scriptsize $\pm 0.366$ & 0.590 \scriptsize $\pm 0.097$ & 0.581 \scriptsize $\pm 0.090$ & 0.581 \scriptsize $\pm 0.090$ & \textbf{0.580} \scriptsize $\pm 0.090$\\
        
        & & F & 0.415 \scriptsize $\pm 0.016$ & \textbf{0.387} \scriptsize $\pm 0.023$ & 0.580 \scriptsize $\pm 0.148$ & 0.465 \scriptsize $\pm 0.049$ & 0.459 \scriptsize $\pm 0.006$ & 0.399 \scriptsize $\pm 0.003$ & 0.395 \scriptsize $\pm 0.006$ & 0.397 \scriptsize $\pm 0.003$\\
        \midrule
        
        \multirow{4}{*}{}USEI & \multirow{2}{*}{}MAE & S &4.681 \scriptsize $\pm 0.564$ & 4.931 \scriptsize $\pm 0.872$ & 4.748 \scriptsize $\pm 0.829$ & 5.244 \scriptsize $\pm  0.939$ & 4.703 \scriptsize $\pm 0.538$ & 4.399 \scriptsize $\pm 0.540$ & 4.318 \scriptsize $\pm 0.533$ & \textbf{4.219} \scriptsize $\pm 0.537$\\
        
        & & F & 11.665 \scriptsize $\pm 0.367$ & 9.454 \scriptsize $\pm 0.809$ & 12.423 \scriptsize $\pm 1.060$ & 21.948 \scriptsize $\pm 0.133$ & 11.118 \scriptsize $\pm 0.220$ & 7.637 \scriptsize $\pm 1.225$ & 7.193 \scriptsize $\pm 1.205$ & \textbf{7.164} \scriptsize $\pm 1.121$\\
        
        & \multirow{2}{*}{}MRE & S & 1.458 \scriptsize $\pm 0.177$ & 1.128 \scriptsize $\pm 0.189$ & 1.088 \scriptsize $\pm 0.162$ & 1.941 \scriptsize $\pm 0.209$ & 1.279 \scriptsize $\pm 0.257$ & 1.201 \scriptsize $\pm 0.214$ & 1.177 \scriptsize $\pm 0.221$ & \textbf{1.031} \scriptsize $\pm 0.193$\\
        
        & & F & 7.473 \scriptsize $\pm 0.623$ & 6.508 \scriptsize $\pm 0.571$ & 8.929 \scriptsize $\pm 2.514$ & 13.706 \scriptsize $\pm 1.268$ & 4.238 \scriptsize $\pm 0.325$ & 2.591 \scriptsize $\pm 0.355$ & \textbf{2.565} \scriptsize $\pm 0.333$ & 2.583 \scriptsize $\pm 0.353$\\
        \midrule
        
        \multirow{4}{*}{}COVID-19 & \multirow{2}{*}{}MAE & S & 7.935 \scriptsize $\pm 0.751$ & 7.144 \scriptsize $\pm 1.159$ & 7.240 \scriptsize $\pm 0.848$ & 7.819 \scriptsize $\pm 1.348$ & 7.566 \scriptsize $\pm 1.095$ & 6.599 \scriptsize $\pm 1.101$ & \textbf{6.427} \scriptsize $\pm 1.043$ & 6.539 \scriptsize $\pm 1.012$\\
        
        & & F & 9.137 \scriptsize $\pm 1.102$ & 9.600 \scriptsize $\pm 1.257$ & 10.409 \scriptsize $\pm 1.910$ & 12.550 \scriptsize $\pm 2.156$ & 9.314 \scriptsize $\pm 0.236$ & 8.696 \scriptsize $\pm 0.426$ & \textbf{8.572} \scriptsize $\pm 0.435$ & 8.635 \scriptsize $\pm 0.447$\\
        
        & \multirow{2}{*}{}MRE & S & 0.564 \scriptsize $\pm 0.126$ & 0.493 \scriptsize $\pm 0.136$ & 0.504 \scriptsize $\pm 0.109$ & 0.769 \scriptsize $\pm 0.169$ & 0.558 \scriptsize $\pm 0.130$ & 0.346 \scriptsize $\pm 0.105$ & \textbf{0.335} \scriptsize $\pm 0.104$ & 0.342 \scriptsize $\pm 0.113$\\
        
        & & F & 0.627 \scriptsize $\pm 0.106$ & 0.556 \scriptsize $\pm 0.052$ & 0.585 \scriptsize $\pm 0.067$ & 0.759 \scriptsize $\pm 0.150$ & 0.585 \scriptsize $\pm 0.007$ & 0.520 \scriptsize $\pm 0.019$ & \textbf{0.507} \scriptsize $\pm 0.017$ & 0.515 \scriptsize $\pm 0.015$\\
    \bottomrule[2pt]
    \end{tabular}
}
\end{table*}

We conducted experiments for both predictive and exploratory analysis to demonstrate the ability of the proposed model in capturing count time sequences with time-varying dynamics. The baseline models included in the experiments are: $\left. 1 \right)$ Gamma process dynamic Poisson factor analysis (GP-DPFA)~\cite{acharya2015nonparametric}. $\left. 2 \right)$ Poisson-gamma dynamical system (PGDS)~\cite{schein2016poisson}. $\left. 3 \right)$ Gamma Markov chains on the rate parameter of gamma distribution (GMC-RATE)~\cite{fevotte2009nonnegative}. $\left. 4 \right)$ Gamma Markov chains on the rate parameter with hierarchical auxiliary variable (GMC-HIER)~\cite{filstroff2021comparative}.  $\left. 5 \right)$ Autogressive beta-gamma procecss (BGAR)~\cite{lewis1989gamma}. BGAR is also a gamma Markov model. We included it as one of the baseline models since a review ~\cite{filstroff2021comparative} highlighted its superiority in well-defined stationary distribution compared with the above-mentioned models.  

The real-world datasets used in the experiments are: $\left. 1 \right)$ \textbf{Integrated Crisis Early Warning System (ICEWS)}~\cite{schein2016poisson}: ICEWS is an international relations event dataset, comprising interaction events between countries extracted from news corpora. For ICEWS dataset, we have $T=365$ time steps and $V=6197$. $\left. 2 \right)$ \textbf{NIPS}~\cite{nips_conference_papers_1987-2015_371}: NIPS dataset contains the papers published in the NeurIPS conference from 1987 to 2015. We have $T=28$ time steps and $V=2000$. $\left. 3 \right)$ \textbf{U.S. Earthquake Intensity (USEI)}~\cite{usgs_earthquake_2025}: USEI contains a collection of damage and felt reports for U.S. (and a few other countries) earthquakes. We use the monthly reports from 1957-1986 and have $T=348, V=64$. $\left. 4 \right)$ \textbf{COVID-19}~\cite{bing_covid19_tracker}: This dataset contains daily death cases data for states in the United States, spanning from March 2020 to June 2020. For this dataset, we have $V=51$ and $T=90$ time steps.

\noindent\textbf{Predictive Analysis.} 
To compare the predictive performance of the proposed model with the baselines, we considered two standard tasks: data smoothing and forecasting. For data smoothing task, our objective is to predict $\boldsymbol{y}^{(t)}$ given the remaining data observation $\boldsymbol{Y} \backslash \boldsymbol{y}^{(t)}$. To this end, we randomly masked 10 percents of the observed data over non-adjacent time steps, and predicted the masked values. For forecasting task, we held out data of the last $S$ time steps, and predicted $\boldsymbol{y}^{(T+1)}, \cdots, \boldsymbol{y}^{(T+S)}$ given $\boldsymbol{y}^{(1)}, \cdots, \boldsymbol{y}^{(T)}$. In this experiment we set $S=2$. We ran the baseline models including GP-DPFA, PGDS, GMC-RATE, GMC-HIER, BGAR, using their default settings as provided in~\cite{acharya2015nonparametric,schein2016poisson,filstroff2021comparative}. For the TV-PGDS, we set $\tau_0=1,\gamma_0=50,\epsilon_0=0.1$. To maintain experimental consistency and ensure comparison, we fixed the number of sub-intervals at $I=6$ for all datasets, with $M$ determined accordingly. Additionally, we set $K=100$ for the ICEWS dataset to capture its higher dimensionality, while $K=10$ was applied to others. These configurations were kept strictly uniform across all baseline models to eliminate any potential bias. While these empirical settings demonstrate robust performance, we further provide a sensitivity analysis of $M$ (or $I$) in the following section to evaluate their influence on model behavior. 

We performed 4000 Gibbs sampling iterations. In the experiments, we found the Gibbs sampler started to converge after 1000 iterations, and thus we set the burn-in time be 2000 iterations. We retained every hundredth sample, and averaged the predictions over the samples. Mean relative error (MRE) and mean absolute error (MAE) are adopted to evaluate the model's predictive capability, which are defined as $\mathrm{MRE} = \frac{1}{TV} \sum_{t}\sum_{v}\frac{\mid y_v^{(t)} - \hat{y}_v^{(t)} \mid}{1 + y_v^{(t)}}$ and $\mathrm{MAE} = \frac{1}{TV}\sum_{t}\sum_{v}\mid y_v^{(t)} - \hat{y}_v^{(t)} \mid$, where $y_v^{(t)}$ indicates the true count and $\hat{y}_v^{(t)}$ is the prediction.

As the experiment results shown in Table \ref{tab:predictive_results}, the TV-PGDS exhibits improved performance in both data smoothing and forecasting tasks. We attribute this enhanced capability to the time-varying transition kernels, which effectively adapt to the non-stationary environment, and thus achieve improved predictive performance. For some datasets (e.g. ICEWS) and tasks, the effectiveness of the Dir-Gam-Dir and PR-Gam-Dir constructions is not evident in the numerical results. This indicates that the selection of these three Dirichlet Markov constructions should depend on different data features. However, the Dir-Gam-Dir and PR-Gam-Dir constructions indeed induce more informative patterns compared with Dir-Dir construction, as shown in the exploratory analysis. 
\begin{figure*}
    \vspace{-1em}
    \centering
    \includegraphics[width=0.9\linewidth]{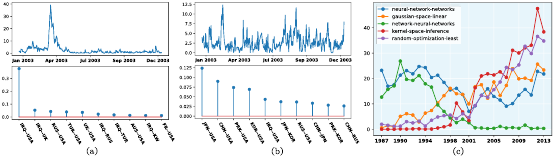}
    \vspace{-1em}
    \caption{The latent factors inferred by the TV-PGDS. (a) and (b) illustrate the top 2 latent factors inferred from ICEWS dataset, (a) corresponds to Iraq war and (b) corresponds to the Six-Party Talks. (c) illustrates the evolving trends of the top 5 latent factors inferred from NIPS dataset.}
    \label{fig:inferred_topics}
\end{figure*}

\noindent\textbf{Exploratory Analysis.} 
We used ICEWS and NIPS datasets for exploratory analysis, and chose the TV-PGDS with Dirichlet-Dirichlet Markov chains for illustration. Figure \ref{fig:inferred_topics}(a) and Figure \ref{fig:inferred_topics}(b) demonstrate the top 2 latent factors inferred by TV-PGDS from ICEWS dataset. From Figure \ref{fig:inferred_topics}(a) we can see that the main labels are ``Iraq (IRQ)--United States (USA)", ``Iraq (IRQ)--United Kingdom (UK)", ``Russia (RUS)--United States (USA)", and so on. This latent factor probably corresponds to the topic about Iraq war. Besides, in Figure \ref{fig:inferred_topics}(a), there is a peak around March, 2003, and we know that the Iraq war broke out exactly on 20 March, 2003. In addition, the most dominant labels shown in Figure \ref{fig:inferred_topics}(b) are ``Japan (JPN)--United States (USA)", ``China (CHN)--United States (USA)", ``North Korea (PRK)--United States (USA)", ``South Korea (KOR)--United States (USA)", and so on. We can infer that this latent factor corresponds to ``Six-Party Talks" and other accidents about it.

Figure \ref{fig:inferred_topics}(c) demonstrates the evolving trends of the top 5 latent factors inferred by the TV-PGDS from NIPS dataset, and the legend indicates the representative words of the corresponding latent factors. Clearly, the green and blue lines correspond to the latent factors of neural network research which started to decline from the 1990s. From the 1990s, the latent factors about statistical and probabilistic methods began to dominate the NeurIPS conference. In addition, the TV-PGDS also captured the revival of neural networks (blue line) from the 2010s. The above observations from the latent structure inferred by the TV-PGDS match our prior knowledge. 

\begin{figure}[t]
    \includegraphics[width=0.90\linewidth]{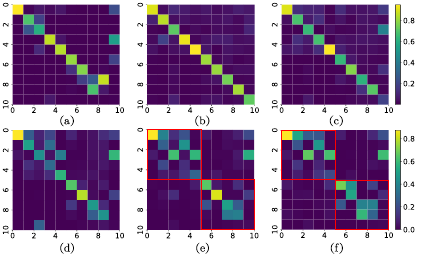}
    \centering
    \caption{Transition matrices inferred from NIPS dataset. (a) The transition matrix inferred by the PGDS. (b)-(f) The time-varying transition matrices inferred by the TV-PGDS.}
    \vspace{-1em}
    \label{fig:nips_matrices}
\end{figure}

\begin{figure}[h]
    \centering
    \includegraphics[width=\linewidth]{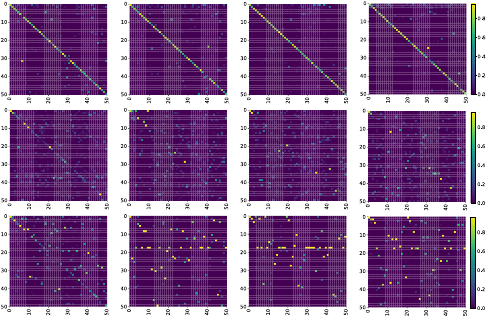}
    \caption{From top to bottom are the first four transition matrices inferred by different Dirichlet Markov chains from ICEWS dataset. Top row: Matrices inferred by the Dir-Dir construction. Middle row: Matrices inferred by the Dir-Gam-Dir construction. Bottom row: Matrices inferred by the PR-Gam-Dir construction.}
    \label{fig:transition_matrices}
\end{figure}
Next, we explored the time-varying transition matrices inferred by the TV-PGDS. We chose NIPS dataset for illustratiuon, and set $K=10$ and the interval length $M$ to be 5. The time-varying transition matrices are shown from Figure \ref{fig:nips_matrices}(b) to Figure \ref{fig:nips_matrices}(f). At the beginning, matrices shown in Figure \ref{fig:nips_matrices}(b) and Figure \ref{fig:nips_matrices}(c) are close to identity matrices. Then the transition matrices tend to become block diagonal matrices with 2 blocks, as shown in Figure \ref{fig:nips_matrices}(d)-\ref{fig:nips_matrices}(f). The representative words for latent factors in the first block are ``state-linear-classification", ``network-neural-networks", ``kernel-image-space", ``network-neural-networks", ``neural-networks-state". The representative words for latent factors in the second block are ``image-sparse-matrix", ``kernel-supervised-random", ``matrix-sample-random", ``inference-prior-latent", ``state-policy-gamma". The first block primarily captured the correlations among the research topics about neural networks. The second block reflects that, from the 1990s, statistical learning and Bayesian methods began to dominate, and these topics are highly correlated. Figure \ref{fig:nips_matrices}(a) illustrates the transition matrix inferred by the PGDS, which is averaged over all time steps. Compared with the TV-PGDS, the PGDS can not capture the informative time-varying transition dynamics.

We also compared the theoretical and empirical properties of the three proposed Dirichlet Markov chain variants. The top row of Figure \ref{fig:transition_matrices} demonstrates transition matrices of the first four sub-intervals of ICEWS dataset inferred by the TV-PGDS (Dir-Dir). Because of the Dir-Dir construction, the consecutive transition matrices smoothly change over time and thus the TV-PGDS may lack sufficient flexibility to capture rapid dynamics.
The middle row of Figure \ref{fig:transition_matrices} illustrates the transition matrices inferred by the TV-PGDS (Dir-Gam-Dir), which takes mutations among latent components into account and captured more complicated patterns. Transition matrices inferred by the PR-Gam-Dir construction are shown in the bottom row of Figure \ref{fig:transition_matrices}, these matrices not only exhibited sufficient flexibility but also captured sparser patterns compared with the Dir-Gam-Dir construction.

\begin{figure}
  \centering
  \includegraphics[width=\linewidth]{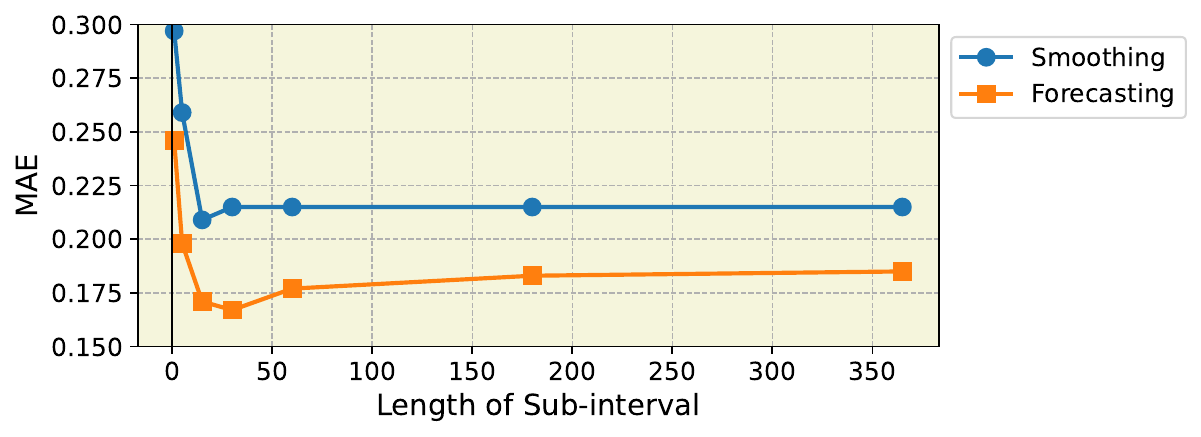}
  \includegraphics[width=\linewidth]{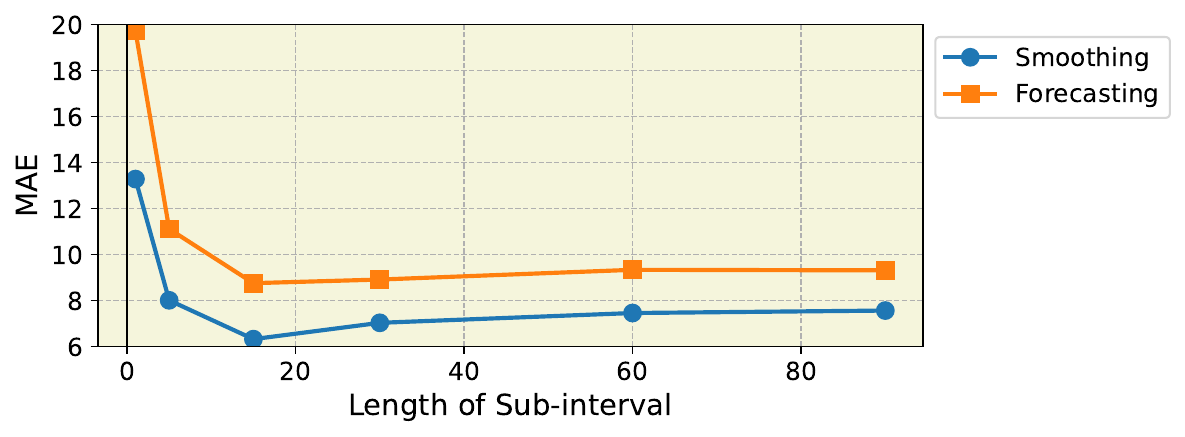}
  \caption{The mean absolute error (MAE) for smoothing and forecasting tasks with different length of sub-interval. Top: the MAE for ICEWS dataset. Down: the MAE for COVID-19 dataset.}
  \label{fig:two-images-column}
\end{figure}

\noindent\textbf{Effect of Sub-interval Length.} 
The length of each sub-interval plays a critical role in the performance of the proposed model. Figure \ref{fig:two-images-column} illustrates, under various sub-interval lengths, the performance of TV-PGDS (Dir-Dir) varies across the ICEWS and COVID-19 datasets. Intuitively, allowing the transition matrix to change at every time step offers greater modeling flexibility compared to dividing the whole interval into fixed-length sub-intervals. A shorter sub-interval length enables the model to more closely track the temporal dynamics of the underlying process. However, shorter sub-intervals also result in fewer data points available for estimating the transition matrices within each segment, which increases estimation variance and may lead to degraded performance. Conversely, longer sub-intervals provide more data for reliable parameter estimation, thereby reducing variance. Nonetheless, overly long sub-intervals may hinder the model’s ability to capture fine-grained temporal variations, as the assumption of a fixed transition matrix over an extended period becomes less realistic. We believe that incorporating a change-point detection mechanism—which adaptively partitions the entire time horizon into irregular sub-intervals based on shifts in transition dynamics—would further enhance the model’s flexibility and accuracy. We leave this as our future work.

\section{Conclusion}
Poisson-gamma dynamical systems with time-varying transition matrices, have been proposed to capture time-varying dynamics observed in count sequences. Dirichlet Markov chains are constructed to allow the underlying transition matrices to evolve over time. Although the Dirichlet Markov processes lack conjugacy, we have developed tractable-but-efficient Gibbs sampling algorithms to perform posterior simulation. For future work, we plan to design methods that can find the point of change, and thus the length of each sub-interval can be determined automatically. 
We also consider to capture the time-varying dynamics  occurring in temporal social networks~\cite{AAAI18,ICML18,ICDM25,UAI26} and irregularly-observed social event data~\cite{UAI20,SDM23,AAAI24}.

\section*{acknowledgements}

Sikun Yang is supported by National Natural Science Foundation of China(NSFC)(Grant No.62476047), Peking University Mathematics Challenge Funding Program(Grant No.2024SRMC10), the Guangdong-Dongguan Joint Research Fund(Grant No.2025A1515140098), and Dongguan Key Laboratory for AI and Dynamical Systems, Dongguan Key Laboratory for Intelligence and Information Technology, Dongguan Key Laboratory for Data Science and Intelligent Medicine, Guangdong Provincial Key Laboratory of Mathematical and Neural Dynamical Systems (Grant No.2024B1212010004), and Guangdong Multidisciplinary Innovation Research Group for New-Generation Intelligent Systems and Diagnostic-Therapeutic Applications(Grant No.2025KCXTD031). 

\bibliographystyle{IEEEtran}
\bibliography{tv_pgds}

\section{Appendix A: MCMC Inference}

\noindent\textbf{Negative Binomial Distribution}  
Let $y \sim \mathrm{Pois} \left( c \lambda \right)$, and $\lambda \sim \mathrm{Gam} \left( a, b\right)$. If we marginalize over $\lambda$, then $y \sim \mathrm{NB} \left( a, \frac{c}{b+c} \right)$. We can further parameterize it as $y \sim \mathrm{NB} \left( a, g \left( \zeta \right) \right)$, where $g\left( z \right) = 1 - \mathrm{exp} \left( -z \right)$ and $\zeta = \mathrm{ln} \left( 1 + \frac{c}{b} \right)$.

\noindent\textbf{Backward decoupling $\theta_k^{(t)}$:} Following the backward decoupling strategy of ~\cite{schein2016poisson}, we recursively marginalize the latent states $\theta_k^{(t)}$ backward in time to decouple the Markov chain. Defining $y_{\cdot k}^{(t)}=\sum_{v=1}^{V}y_{vk}^{(t)}\sim Pois(\delta^{(t)}\theta_k^{(t)})$, we start at the boundary $t=T$ to marginalize $\theta_k^{(T)}$ and obtain $y_{\cdot k}^{(T)}\sim NB(\tau_0\sum_{k_2=1}^{K}\pi_{kk_2}^{i(T-1)}\theta_{k_2}^{(T-1)}, g(\zeta^{(T)}))$, where $\zeta^{(T)}=\ln(1+\delta^{(T)}/\tau_0)$. 

To proceed backward, we introduce auxiliary variables $l_k^{(T)} \sim CRT(y_{\cdot k}^{(T)}, \tau_0\sum_{k_2=1}^{K}\pi_{kk_2}^{i(T-1)}\theta_{k_2}^{(T-1)})$.By Lemma 1, the joint distribution of $y_{\cdot k}^{(T)}$ and $l_k^{(T)}$ can be expressed as $y_{\cdot k}^{(T)} \sim \mathrm{SumLog}(l_k^{(T)}, g(\zeta^{(T)}))$, $l_k^{(T)}  \sim \mathrm{Pois}\zeta^{(T)} \tau_0 \sum_{k_2=1}^K \pi_{kk_2}^{i\left( T-1 \right)} \theta_{k_2}^{(T-1)} )$. Next, note that $y_{\cdot k}^{(T-1)} \sim \mathrm{Pois} ( \delta^{(T-1)} \theta_k^{(T-1)} )$, if we introduce $m_k^{(T-1)} = y_{\cdot k}^{(T-1)} + l_{\cdot k}^{(T)}$, then we have $m_k^{(T-1)} \sim \mathrm{Pois} ( \theta_k^{(T-1)} ( \delta^{(T-1)} + \zeta^{(T)} \tau_0 ) )$. We can again marginalize over $\theta_k ^{(T-1)}$ to obtain $m_k^{(T-1)} \sim \mathrm{NB} ( \tau_0 \sum_{k_2=1}^K \pi_{kk_2}^{i(T-2)} \theta_{k_2}^{(T-2)} , g ( \zeta^{(T-1)} ))$, where $\zeta^{(T-1)} = \mathrm{ln} ( 1 + \frac{\delta^{(T-1)}}{\tau_0} + \zeta^{(T)} )$. Then we introduce auxiliary variables $l_{k}^{(T-1)} \sim \mathrm{CRT} ( m_k^{(T-1)},  \tau_0 \sum_{k_2=1}^K \pi_{kk_2}^{i(T-2)} \theta_{k_2}^{(T-2)} )$. 

Repeating this marginalization recursively down to $t=1$ with $\zeta^{(t)}=\ln(1+\delta^{(t)}/\tau_0+\zeta^{(t+1)})$ to maginalize over all the $\theta_k^{(t)}$, we decouple the sequence. To sample $\theta_k^{(t)}$, we set $l_{\cdot k}^{(T+1)}=0$ and $\zeta^{(T+1)}=0$, and first sample the auxiliary variables $l_{k\cdot}^{(t)}$ and $l_{kk_2}^{(t)}$ backward for $t=T,\dots,2$ as Eq. (\ref{sampling_l}).

And using the relationship between Poisson and multinomial distributions, we have
\begin{equation}
    \left( l_{1k}^{(t)}, \cdots, l_{Kk}^{(t)} \right) \sim \mathrm{Mult}\left(l_{\cdot k}^{(t)}, \pi_{1k}^{i\left( t-1 \right)},\cdots,\pi_{Kk}^{i\left( t-1 \right)}\right). \label{l_kk_mult}
\end{equation}

\noindent\textbf{Sampling $\boldsymbol{\Pi}^{(i)}$ :} For $i=I$, by Eq.(\ref{l_kk_mult}), $\left( l_{1k}^{(I)}, \cdots, l_{Kk}^{(I)}\right)$ is multinomial distributed. Thus by multinomial-Dirichlet conjugacy, we obtain
\begin{equation}
\label{pi_I}
    \left( \boldsymbol{\pi}_k^{(I)} \mid -\right) \sim \mathrm{Dir}\left( \alpha_{1k}^{(I)} + l_{1k}^{(I)}, \cdots, \alpha_{Kk}^{(I)} + l_{Kk}^{(I)} \right).\notag
\end{equation}

\noindent \textbf{Inference for Dirichlet-Dirichlet Markov chains.} 
For Dirichlet-Dirichlet Markov chains, $\alpha_{k_1k}^{\left(i \right)} = \eta K \pi_{k_1k}^{\left(i-1\right)}$. If we marginalize $\left( \pi_{1k}^{(I)}, \cdots, \pi_{Kk}^{(I)} \right)$, $\left( l_{1k}^{(I)}, \cdots, l_{Kk}^{(I)}\right)$ will be Dirichlet-multinomial distributed as 
\begin{equation}
\resizebox{\linewidth}{!}{$
    \left( l_{1k}^{(I)},\cdots,l_{Kk}^{(I)}\right) \sim \mathrm{DirMult} \left(l_{k\cdot}^{\left(I\right)},\left(\eta K \pi_{1k}^{\left(I-1\right)},\cdots,\eta K \pi_{Kk}^{\left(I-1\right)}\right)\right).\notag
    $}
\end{equation}
Thus by Lemma 2, for $i=I$, we first sample the auxiliary variables as
\begin{align}
    \left( q_k^{(I)} \mid -\right) &\sim \mathrm{Beta} \left( l_{\cdot k}^{(I)}, \eta K\right),\notag\\
    \left(h_{k_1k}^{(I)} \mid - \right) &\sim \mathrm{CRT} \left( l_{k_1k}^{(I)}, \eta K \pi_{k_1k}^{\left(I-1\right)}\right).\notag
\end{align}
Via Lemma 1 and 2, $h_{\cdot k}^{\left(i\right)}$ is Poisson distributed and ${\left( h_{1k}^{(i)},\cdots, h_{Kk}^{(i)} \right)}$ is Dirichlet-multinomial distributed. Thus for $i=I-1,\cdots,2$, we sample the auxiliary variables as
\begin{align}
    \left( q_k^{(i)} \mid -\right) &\sim \mathrm{Beta} \left( l_{\cdot k}^{(i)} + h_{\cdot k}^{(i+1)}, \eta K\right),\notag \\
    \left(h_{k_1k}^{(i)} \mid - \right) &\sim \mathrm{CRT} \left( l_{k_1k}^{(i)} + h_{k_1 k}^{(i+1)}, \eta K \pi_{k_1k}^{\left(i-1\right)}\right), \notag
\end{align}
where $l_{k_1k}^{(i)} = \sum_{\left( i-1 \right)M+1}^{iM}l_{k_1k}^{(t)}$ refers to the summation of $l_{k_1k}^{(t)}$ over $i$-th interval. Via Lemma 2, conditioning on $q_k^{(i)}$, we have
\begin{equation}
    \left( l_{k_1k}^{(i)} + h_{k_1 k}^{(i+1)}\right) \sim \mathrm{NB} \left( \eta K \pi_{k_1k}^{\left(i-1\right)}, q_k^{(i)} \right). \nonumber
\end{equation}

Then via Lemma 1, $\left( l_{k_1k}^{(i)} + h_{k_1 k}^{(i+1)}\right)$ also follows a Poisson distribution, $\left( l_{k}^{(i)} + h_{k}^{(i+1)}\right)$ thus is mutinomial distributed. Via Dirichlet-multinomial conjugacy, for $i=I-1, \cdots, 2$, we obtain
\begin{align}
    \left( \boldsymbol{\pi}_k^{(i)} \mid -\right) \sim \mathrm{Dir} \Big( &  \eta K \pi_{1k}^{(i-1)} + l_{1k}^{(i)} + h_{1k}^{\left(i+ 1\right)}\notag\\ &, \cdots,  \eta K \pi_{Kk}^{(i-1)} + l_{Kk}^{(i)} + h_{Kk}^{\left(i+1 \right)}\Big). \notag
\end{align}
Specifically, for $i=1$, we have
\begin{align}
    \left( \boldsymbol{\pi}_k^{(1)} \mid - \right) \sim \mathrm{Dir} ( &\nu_1 \nu_k + l_{1k}^{(1)} + h_{1k}^{(2)}, \cdots, \xi \nu_k + l_{kk}^{(1)} + h_{kk}^{(2)}\notag\\&, \cdots,\nu_K \nu_k + l_{Kk}^{(1)} + h_{Kk}^{(2)} ). \notag
\end{align}

For sampling $\eta$, note that
\begin{equation}
    \left( h_{k_1k}^{(i)} \mid - \right) \sim \mathrm{Pois} \left( -\eta K \pi_{k_1k}^{(i-1)} \mathrm{ln} \left( 1 - q_k^{(i)} \right) \right), i=I, \cdots, 2. \notag
\end{equation}
 Given the prior $\eta \sim \mathrm{Gam} \left( e_0, f_0 \right)$,via Poisson-gamma conjugacy, we obtain
\begin{align}
    (\eta \mid - ) \sim \mathrm{Gam} (&e_0 + \sum_{i=2}^I \sum_{k_1=1}^K \sum_{k_2=1}^K h_{k_1k_2}^{(i)},\notag\\ & f_0 - K \sum_{i=2}^I \sum_{k=1}^K \mathrm{ln} ( 1 - q_k^{(i)} )).\notag
\end{align}

\noindent \textbf{Inference for Dirichlet-Gamma-Dirichlet Markov chains.} 

$\left( l_{1k}^{(i)}, \cdots, l_{Kk}^{(i)}\right)$ will be Dirichlet-multinomial distributed when marginalizing $\left( \pi_{1k}^{(i)}, \cdots, \pi_{Kk}^{(i)} \right)$. Thus by Lemma 2, for $i=I$, we first sample the auxiliary variables as
\begin{equation}
     \left( q_k^{(I)} \mid -\right) \sim \mathrm{Beta} \left( l_{\cdot k}^{(I)}, \alpha_{\cdot k}^{(I)}\right),\notag
\end{equation}
\begin{equation}
    \left(h_{k_1k}^{(I)} \mid - \right) \sim \mathrm{CRT} \left( l_{k_1k}^{(I)}, \alpha_{k_1k}^{(I)}\right). \notag
\end{equation}
Similarly, by Eq.(\ref{g_mult_gam}), we construct $\left( g_{\cdot 1k}^{(i)},\cdots, g_{\cdot Kk}^{(i)} \right)$ which is also Dirichlet-multinomial distributed. Thus for $i=I-1,\cdots,2$, we sample the auxiliary variables as
\begin{align}
    \left( q_k^{(i)} \mid -\right) &\sim \mathrm{Beta} \left( l_{\cdot k}^{(i)} + g_{\cdot k}^{(i+1)}, \alpha_{\cdot k}^{(i)}\right),\notag\\ \left(h_{k_1k}^{(i)} \mid - \right) &\sim \mathrm{CRT} \left( l_{k_1k}^{(i)} + g_{\cdot k_1 k}^{(i+1)}, \alpha_{k_1k}^{(i)}\right). \notag
\end{align}
Via Lemma 2, conditioning on $q_k^{(i)}$, we have $\left( l_{k_1k}^{(i)} + g_{\cdot k_1 k}^{(i+1)}\right) \sim \mathrm{NB} \left( \alpha_{k_1k}^{(i)}, q_k^{(i)} \right)$. Then via Lemma 1, we obtain $h_{k_1k}^{(i)} \sim \mathrm{Pois} \left( -\alpha_{k_1k}^{(i)} \mathrm{ln} \left( 1 - q_k^{(i)} \right) \right)$. Thus via Poisson-gamma conjugacy, we obtain
\begin{align}
    ( \alpha_{k_1k}^{(i)} \mid - ) \sim \mathrm{Gam} ( &\gamma_k^{(i-1)} \sum_{k2=1}^K \psi^{(i-1)}_{kk_1k_2}\pi_{k_2k}^{(i-1)} + h_{k_1k}^{(i)},\notag\\&c_k^{(i)}-\mathrm{ln}( 1-q_k^{(i)})).\notag
\end{align}
Marginalizing over $\alpha_{k_1k}^{(i)}$, and via the definition of negative binomial distribution, we have
\begin{equation}
    h_{k_1k}^{(i)} \sim \mathrm{NB} \left( \gamma_k^{(i-1)} \sum_{k2=1}^K \psi^{(i-1)}_{kk_1k_2}\pi_{k_2k}^{(i-1)}, \frac{-\mathrm{ln}\left(1-q_k^{(i)} \right)}{c_k^{(i)}-\mathrm{ln}\left( 1-q_k^{(i)}\right)}\right). \nonumber
\end{equation}
Then using Lemma 1, we sample
\begin{equation}
    \left(g_{k_1k}^{(i)}\mid -\right) \sim \mathrm{CRT} \left( h_{k_1k}^{(i)}, \gamma_k^{(i-1)} \sum_{k2=1}^K \psi^{(i-1)}_{kk_1k_2}\pi_{k_2k}^{(i-1)}\right),\notag
\end{equation}
and obtain $g_{k_1k}^{(i)} \sim \mathrm{Pois} (\gamma_k^{(i-1)} \sum_{k2=1}^K \psi^{(i-1)}_{kk_1k_2}\pi_{k_2k}^{(i-1)} \mathrm{ln} (1 - \mathrm{ln}( 1-q_k^{(i)}) \big/ {c_k^{(i)}} ) )$. Using the relationship between Poisson and multinomial distributions and $\sum_{k_1}^K \psi_{kk_1k_2}^{(i-1)}=1$, we have
\begin{align}
    \left( g_{\cdot 1k}^{(i)},\cdots, g_{\cdot Kk}^{(i)} \right) &\sim \mathrm{Mult} \left( g_{\cdot k}^{(i)}, \left( \pi_{k_1k}^{(i-1)} \right)_{k_1=1}^K \right),\label{g_mult_gam} \\
    \left( g_{1k_2k}^{(i)}, \cdots, g_{Kk_2k}^{(i)} \right) &\sim \mathrm{Mult} \left( g_{\cdot k_2k}^{(i)}, \left( \psi_{kk_1k_2}^{(i-1)} \right)_{k1=1}^{K} \right). \notag
\end{align}
Thus by Dirichlet-multinomial conjugacy, for $i=I,\cdots,2$, we can obtain the posterior distributions of $( \psi_{k1k_2}^{(i-1)}, \cdots, \psi_{kKk_2}^{(i-1)} )$ and $ \boldsymbol{\pi}_k^{(i-1)}$, which is the same as in Eq.(\ref{sampling_psi}) and (\ref{sampling_pi}).
For sampling $\gamma_k^{(i-1)}$, note that $\sum_{k_1}^K \psi_{kk_1k_2}^{(i-1)}=1$, we have
\begin{align}
    &g_{\cdot k}^{(i)} \sim \mathrm{Pois} \left( \gamma_k^{(i-1)} \mathrm{ln} \left(1 - \mathrm{ln}\left( 1-q_k^{(i)}\right) \big/ {c_k^{(i)}} \right) \right).\notag
\end{align}
Thus via Poisson-gamma conjugacy, we obtain
\begin{equation}
    \left(\gamma_k^{(i-1)}\mid-\right) \sim \mathrm{Gam} \left( \epsilon_0 + g_{\cdot k}^{(i)},  \epsilon_0 + \mathrm{ln} \left(1 - \mathrm{ln}\left( 1-q_k^{(i)}\right) \right) \right).\notag
\end{equation}
By gamma-gamma conjugacy, we have
\begin{equation}
    \left( c_k^{(i)}\mid -\right) \sim \mathrm{Gam} \left(\epsilon_0 + \gamma_k^{(i-1)}, \epsilon_0 + \sum_{k_1=1}^K \alpha_{k_1k}^{(i)}\right).\notag
\end{equation}

\noindent \textbf{Sampling $\nu_k$ and $\xi$ :} As we sample $\boldsymbol{\Pi}^{(i)}$, by the definition of Dirichlet-multinomial distribution, we obtain
\begin{align}
   &\big( l_{1k}^{(1)}+g_{\cdot 1k}^{(2)}, \cdots, l_{Kk}^{(1)}+g_{\cdot Kk}^{(2)}\big) \notag\\ &\sim \mathrm{DirMult} \left( \nu_1 \nu_K, \cdots, \xi \nu_k, \cdots, \nu_K \nu_k \right). \notag
\end{align}
In particular, with a little abuse of notation here, for Dir-Dir construction, we take $g_{\cdot k_1k}^{(2)}=h_{k_1k}^{(2)}$.
We first sample 
\begin{equation}
    \left(h_{k_1k}^{(1)} \mid - \right) \sim \left\{
    \begin{array}{cc}
         \mathrm{CRT} \left( l_{k_1k}^{(1)}+g_{\cdot k_1k}^{(2)}, \nu_{k_1} \nu_k \right) &  k_1 \neq k\\
         \mathrm{CRT} \left( l_{k_1k}^{(1)}+g_{\cdot k_1k}^{(2)}, \xi \nu_k \right)\quad & k_1 = k. \notag
    \end{array} \right.
\end{equation}
Then we sample
\begin{equation}
\label{q_1}
    q_k^{(1)} \sim \mathrm{Beta} \left( l_{\cdot k}^{(1)}+g_{\cdot k}^{(2)}, \nu_k \left( \sum_{k_1 \neq k} \nu_{k1} + \xi \right) \right). \notag
\end{equation}
We further introduce
\begin{align}
    n_k &=  h_{kk}^{(1)} + \sum_{k_1 \neq k} h_{k_1k}^{(1)} + \sum_{k_2 \neq k} h_{kk_2}^{(1)} + l_{k\cdot}^{(1)}, \notag\\
    \rho_k &= \tau_0 \zeta^{(1)} -\mathrm{ln} \left( 1 - q_k^{(1)} \right) \left( \xi + \sum_{k_1 \neq k} \nu_{k_1} \right) \notag\\&\quad- \sum_{k_2 \neq k} \mathrm{ln} ( 1 - q_{k_2}^{(1)} ) \nu_{k_2}. \notag
\end{align}
Via Poisson-gamma conjugacy, we have
\begin{align}
    \left( \xi \mid - \right)  &\sim \mathrm{Gam} \left( \frac{\gamma_0}{K} + \sum_k h_{kk}^{(1)}, \beta - \sum_{k} \nu_k \mathrm{ln} \left( 1 - q_k^{(1)} \right)\right), \notag\\
    \left( \nu_k \mid - \right)  &\sim \mathrm{Gam} \left( \frac{\gamma_0}{K} + n_k, \beta + \rho_k \right). \notag
\end{align}

\noindent \textbf{Sampling $\delta^{(t)}$ and $\beta$ :} Via Poisson-gamma conjugacy
\begin{equation}
    \left( \delta^{(t)} \mid - \right) \sim \mathrm{Gam} \left( \epsilon_0 + \sum_{v=1}^V y_v^{(t)}, \epsilon_0 + \sum_{k=1}^K \theta_k^{(t)} \right).\notag
\end{equation}
And by gamma-gamma conjugacy, we obtain
\begin{equation}
    \left( \beta \mid -\right) \sim \mathrm{Gam} \left( \epsilon_0 + \gamma_0, \epsilon_0 + \sum_{k=1}^K \nu_k \right).\notag
\end{equation}

\end{document}